\documentclass[10pt,twocolumn,letterpaper]{article}

\usepackage{wacv}
\usepackage{times}
\usepackage{epsfig}
\usepackage{graphicx}
\usepackage{xcolor}
\usepackage{amsmath}
\usepackage{amssymb}
\usepackage{pifont}
\usepackage{lipsum}  
\usepackage{algorithm}
\usepackage{booktabs}
\usepackage{algpseudocode}
\usepackage{tikz}
\usepackage{caption}
\usepackage{subcaption}
\usepackage{enumitem}
\usepackage{wrapfig}

\newcommand*\circled[1]{\tikz[baseline=(char.base)]{
            \node[shape=circle,draw,inner sep=1pt] (char) {#1};}}

\usepackage{multirow}
\usepackage{pifont}

\newcommand{\cmark}{\ding{51}}%
\newcommand{\xmark}{\ding{55}}%

\newcommand{\minisection}[1]{\vspace{0.03in} \noindent {\bf #1}\ \ }

\newsavebox\CBox
\newlength\CLength
\def\numcircledpict#1{\sbox\CBox{#1}%
  \ifdim\wd\CBox>\ht\CBox \CLength=\wd\CBox\else\CLength=\ht\CBox\fi
    \makebox[1.5\CLength]{\makebox(0,1.5\CLength){\put(0,0){\circle{1.5\CLength}}}%
    \makebox(0,1.5\CLength){\put(-.5\wd\CBox,0){#1}}}}

\usepackage[utf8]{inputenc} 
\usepackage[T1]{fontenc}    
\usepackage[pagebackref,breaklinks,colorlinks]{hyperref}
\usepackage{url}            
\usepackage{booktabs}       
\usepackage{amsfonts}       
\usepackage{nicefrac}       
\usepackage{microtype}      
\usepackage{xcolor}         

\def\wacvPaperID{0055} 

\wacvalgorithmstrack   
\wacvfinalcopy

\title{Plasticity-Optimized Complementary Networks for Unsupervised Continual Learning}

\author{Alex Gomez-Villa$^{1,2}$, Bartlomiej Twardowski$^{1,2,3}$, Kai Wang$^{1,2}$\thanks{Corresponding author}, Joost van de Weijer$^{1,2}$\\
$^1$Computer Vision Center, Barcelona, Spain\\$^2$ Universitat Autonoma de Barcelona, Barcelona, Spain\\
$^3$ IDEAS NCBR, Warsaw, Poland\\
{\tt\small \{agomezvi,btwardowski,kwang,joost\}@cvc.uab.es}
}

\begin{document}

\maketitle

\begin{abstract} 
 
Continuous unsupervised representation learning (CURL) research has greatly benefited from improvements in self-supervised learning (SSL) techniques.  As a result, existing CURL methods using SSL can learn high-quality representations without any labels, but with a notable performance drop when learning on a many-tasks data stream.
We hypothesize that this is caused by the regularization losses that are imposed to prevent forgetting, leading to a suboptimal plasticity-stability trade-off: they either do not adapt fully to the incoming data (low plasticity), or incur significant forgetting when allowed to fully adapt to a new SSL pretext-task (low stability). In this work, we propose to train an expert network that is relieved of the duty of keeping the previous knowledge and can focus on performing optimally on the new tasks (optimizing plasticity). In the second phase, we combine this new knowledge with the previous network in an adaptation-retrospection phase to avoid forgetting and initialize a new expert with the knowledge of the old network. We perform several experiments showing that our proposed approach outperforms other CURL exemplar-free methods in few- and many-task split settings. Furthermore, we show how to adapt our approach to semi-supervised continual learning (Semi-SCL) and show that we surpass the accuracy of other exemplar-free Semi-SCL methods and reach the results of some others that use exemplars.
\end{abstract}

\section{Introduction}

Continual learning (CL) designs algorithms that can learn from shifting distributions (non-IID data), generally this is modeled by learning from a sequence of tasks~\cite{delange2021continual}. The main challenge for these methods is the problem of catastrophic forgetting~\cite{mccloskey1989catastrophic}, which is a dramatic drop in performance on previous tasks. Most CL approaches, therefore, need to address the trade-off between acquiring new knowledge (plasticity) and preventing forgetting of previous knowledge (stability). The vast majority of existing methods in continual learning have focussed on supervised learning, where the incoming data stream is fully labeled. In this paper, we focus on continual learning on unsupervised data. 

Only recently, some works have explored continual learning on 
 unsupervised non-IID data-streams~\cite{gomez2022continually,fini2022self}. Motivated by the tremendous progress in unsupervised learning, notably of contrastive learning approaches~\cite{chen2020SimCLR,zbontar2021barlow}, methods aim to extend these methods to the continual setting. An additional motivation is the fact that unsupervised learning tends to lead to more generalizable feature representations since features that are not relevant to the specific discriminative task are not automatically discarded. This can potentially lead to representations that can faster incorporate new tasks without incurring significant amounts of forgetting. PFR~\cite{gomez2022continually} uses a projection head after the feature extractor of an SSL framework to predict past representations. Projected representations are motivated to be close to past representations; therefore, the present model is encouraged to remember past knowledge. CaSSLe~\cite{fini2022self} uses a similar strategy as PFR, but the projection head is used after the projector of the SSL approach. Even though these methods obtain satisfactory results, they struggle to adapt to new tasks without jeopardizing the vast knowledge already accumulated by the network. We hypothesize that the regularization imposed by these methods to avoid forgetting hurts the learning process of CURL in the following ways: 1) the SSL component cannot fully adapt to the incoming data (low plasticity), 2) The model will have a significant drift (forgetting) when the current model is unable to perform the SSL pretext-task. These effects increase as the number of tasks increases, and consequently, the data for training reduces.

Complementary learning systems (CLS) theory~\cite{mcclelland1995there,kumaran2016learning} proposes a computational framework in which the interplay between a fast (episodic memory/specific experience) and a slow (semantic/general structured) memory system is the core of the mechanism of knowledge consolidation. Several existing CL methods have taken inspiration from CLS as a way to find a good stability-plasticity trade-off (see~\cite{o2002hippocampal,parisi2019continual} for a review). 
The fast learner can quickly adapt to new knowledge, which then is carefully absorbed by the slower learner. DualNet~\cite{pham2021dualnet} proposes to use a self-supervised method to train the slow, more generic learner, which weights can be quickly adapted for solving a supervised task with exemplars from the replay buffer. Recently, in~\cite{arani2022learning} the authors proposed to use a CLS-based approach and maintain separate plastic and stable models for online CL with experience replay. However, existing methods that exploit CLS for continual learning have in common that they only consider a supervised learning scenario.

This paper aims to apply complementary learning systems theory to improve continual learning from unsupervised data streams. The existing methods~\cite{gomez2022continually,fini2022self} can suffer from suboptimal stability and plasticity on longer sequences since they have difficulty adapting to the new knowledge required to address the latest task, while maintaining the vast knowledge already learned on earlier tasks. Instead, we propose to train an expert network that is relieved of the task of keeping the previous knowledge and can focus on the task of performing optimally on new tasks. In the second phase, we combine this new knowledge with the old network in an adaptation-retrospection phase to avoid forgetting.
In conclusion, the main contributions of this work are : 
\begin{itemize}

\item A new exemplar-free continual unsupervised representation learning (CURL) method called \textit{Plasticity-Optimized COmplementary Networks} (POCON). Existing CURL methods learn new knowledge while imposing regularization to prevent forgetting. Instead, POCON separates the learning of new knowledge from the knowledge integration part. Analysis confirms that this leads to a better stability-plasticity trade-off.
\item Extensive experiments confirm that POCON outperforms state-of-the-art CURL on various settings (e.g., a 5-9 \% performance gain over CaSSLe on ImageNet100 for a 20-100 task-split). Unlike previous CURL methods, POCON can thrive in low-data regimes (such as small-task incremental learning) and setups without task boundaries. We also demonstrate the application of POCON to semi-supervised continual learning. 
\item We propose and evaluate a 
\emph{heterogeneous} version of POCON, where the main network can have a different network architecture than the expert. This opens up the possibility for interesting applications where a slow/big network can be deployed in a cloud environment, while a fast/slow learner can be utilized on an edge device, such as a mobile phone.
    
\end{itemize}

\section{Related work}

\paragraph{Continual Learning and Class Incremental Learning.} Existing continual learning methods can be broadly categorized into three types: replay-based, architecture-based, and regularization-based methods~\cite{delange2021continual, masana2022class}. Replay-based methods either save a small amount of data from previously seen tasks~\cite{bang2021rainbow, chaudhry2019tiny} or generate synthetic data with a generative model~\cite{wang2021ordisco, zhai2021hyper}. The replay data can be used during training together with the current data, such as in iCaRL~\cite{rebuffi2017icarl} and LUCIR~\cite{hou2019learning}, or to constrain the gradient direction while training, such as in AGEM~\cite{chaudhry2018efficient}. Architecture-based methods activate different subsets of network parameters for different tasks by allowing model parameters to grow with the number of tasks. Previous works following this strategy include HAT~\cite{serra2018overcoming}, Piggyback~\cite{mallya2018piggyback}, PackNet~\cite{mallya2018packnet}, DER~\cite{yan2021dynamically} and Ternary Masks~\cite{masana2020ternary}. Regularization-based methods add a regularization term derived from knowledge of previous tasks to the training loss. This can be done by either regularizing the weight space, which constrains important parameters~\cite{shi2021continual,tang2021layerwise}, or the functional space, which constrains predictions or intermediate features~\cite{douillard2020podnet, cheraghian2021semantic, hu2021distilling}. EWC~\cite{kirkpatrick2017overcoming}, MAS~\cite{aljundi2018memory}, REWC~\cite{liu2018rotate}, SI~\cite{zenke2017continual}, and RWalk~\cite{chaudhry2018riemannian} constrain the importance of network parameters to prevent forgetting. Methods such as LwF~\cite{li2017learning}, LwM~\cite{dhar2019learning}, and BiC~\cite{wu2019large} leverage knowledge distillation to regularize features or predictions. DMC~\cite{zhang2020class} work is more related to POCON as the authors proposed to train the expert network without any regularization for a classification task. However, after that, in a distillation phase an additional auxiliary dataset is used to integrate old and new knowledge.

\minisection{Self-supervised representation learning.}
In recent years, unsupervised methods based on self-supervision have become dominant in learning representation for computer vision systems. The aim of self-supervised learning is to acquire high-quality image representations without explicit labeling. Initially, these methods addressed some well-defined pretext tasks, such as predicting rotation~\cite{gidaris2018_pretext_rotation}, determining patch position~\cite{doersch2015_pretext_patches}, or solving jigsaw puzzles in images~\cite{noroozi2016_pretext_puzzle}, and labels for these discriminative pretext tasks can be automatically computed to enable learning of meaningful feature representations of images. Recently, researchers have adapted contrastive methods for use with unlabeled data and have placed more emphasis on instance-level data augmentation to find similar or contrasting samples~\cite{bardes2022vicreg,caron2020SwAV, chen2020SimCLR, NEURIPS2020BYOL,zbontar2021barlow}. These methods heavily rely on stochastic data augmentation~\cite{xiao2021should, zini2023planckian} to generate sufficient similar examples for learning representations, and negative examples are either randomly sampled or excluded entirely~\cite{chen2021SimSiam}. 

 Self-supervised learning has also been used to improve the learning of a sequence of supervised tasks~\cite{zbontar2021barlow,zhu2021prototype}. Their objective is not to learn from unlabeled data, but rather to use self-supervised learning to further enrich the feature representation. Similarly, pre-trained models with self-supervision have been used to improve incremental average classification metrics~\cite{gallardo2021self} with data augmentation, distillation, and even exemplars. Training self-supervised models directly on class-IL setting without exemplars was also proposed in~\cite{gomez2022continually, fini2022self}, where both present results that self-supervised learning mitigates the problem of forgetting.

 \begin{table}[]
 \centering

\begin{tabular}{l|cccc}
\midrule
Method      & \emph{Labels} & \emph{Exemplars} & \emph{Regularization} \\
\midrule
DMC~\cite{zhang2020class}     & \checkmark      &  \checkmark$^*$    & \ding{55}            \\
DualNet~\cite{pham2021dualnet}     & \checkmark          & \checkmark  & \checkmark        \\
CLS-ER~\cite{arani2022learning}       & \checkmark          & \checkmark & \checkmark           \\
LUMP~\cite{madaan2022representational}        & \ding{55}        & \checkmark    & \checkmark        \\
\midrule
PFR~\cite{gomez2022continually}         &    \ding{55}       & \ding{55}   & \checkmark     \\
CaSSLe~\cite{fini2022self}      & \ding{55}          & \ding{55}   & \checkmark     \\
POCON (ours) & \ding{55}          & \ding{55}  & \ding{55}   \\
\midrule  
\end{tabular}

\caption{Comparison of SSL-based CURL. Only POCON does not use any regularization during the training, what results in higher plasticity for the current task. $^*$DMC uses an auxilary dataset for distillation, instead of exemplars.}
\label{tab:similarMethod}
\end{table}

\minisection{Complementary learning systems.}
There are CLS-based methods that use several networks in addition to a rehearsal memory or pseudo-sample generator. FearNet~\cite{kemker2017fearnet} uses a hippocampal network for recalling new examples, a PFC network for long-term memories, and a third network to select between the PFC or hippocampal networks for a particular instance. In~\cite{parisi2018lifelong}, they propose a G-EM network that performs unsupervised learning from spatiotemporal representations and a G-SM network that uses signals from G-EM and class labels to learn from videos incrementally. Closer to our work are DualNet~\cite{pham2021dualnet} and CLS-ER~\cite{arani2022learning} (previously explained); however, both models are supervised and use exemplars. Table~\ref{tab:similarMethod} presents a summary of similar CL methods.


\section{Method}

In this section, we describe our approach for continual learning of self-supervised representations, referred to as Plasticity-Optimized COmplementary Networks (POCON), which eliminates the need for memory or replay. Our method is based on the complementary learning system (CLS) framework, and involves an interplay between a fast-expert learner and a slow-main network. Our work is motivated by the recognition that fast adaptation to new data is crucial in constructing more robust representations during continual unsupervised representation learning. Rather than attempting to maintain network stability in its old state through strict or relaxed distillation methods, as suggested in recent works~\cite{fini2022self,gomez2022continually}, POCON allows the expert network to learn a new task freely without any restrictions. After acquiring the new knowledge, we integrate it into the main network. In turn, the main network serves as a good starting point for the new expert. Before presenting the details of the POCON method, we first provide a brief introduction to the problem of continual self-supervised representation learning.

\begin{figure*}[tb]
  \centering
  \includegraphics[width=0.9\textwidth]{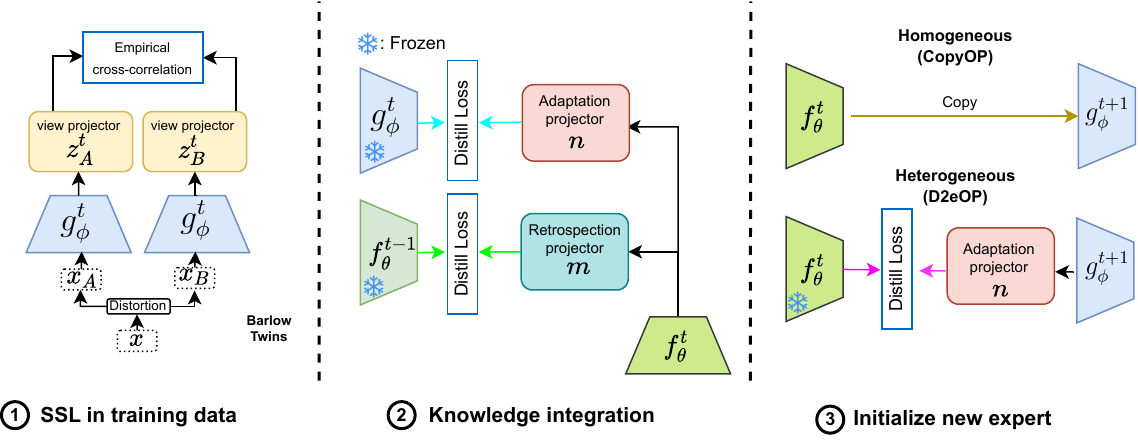}
  \caption{Overview of the POCON method that uses multi-stage training of complementary learning system for unsupervised continual representation learning. Three stages of training allow POCON to maintain fast and efficient adaptation of the fast expert network while maintaining stable representation in a slow learner – the main network. To avoid forgetting, knowledge from the new task expert is integrated (adaptation) with the previous state of the slow network (retrospection). The last stage of the training is dedicated for the best preparation of the new expert for the new task.}
  \label{fig:main_method}
\end{figure*}

\subsection{Self-supervised representation learning}

In recent research on self-supervised learning, the objective is to train a network $f_\theta:\mathcal{X}\rightarrow \mathcal{F}$, which maps input data from $\mathcal{X}$ to output feature representations in $\mathcal{F}$. The network is trained using unlabeled input data $x$ sampled from a distribution $\mathcal{D}$. The learned feature representation is subsequently used to facilitate various downstream tasks. In this paper, we employ the BarlowTwins approach~\cite{zbontar2021barlow} for self-supervised learning of the representation network $g_{\theta}$. This approach serves as a common baseline for previous works~\cite{fini2022self,gomez2022continually}. 
However, the proposed method is versatile and can be extended to other self-supervised techniques.

In Fig.~\ref{fig:main_method} (left) the BarlowTwins architecture is presented. Both branches use a projector network $z$ 
and both share the same parameters, but while computing the empirical cross-correlation loss operates on different views of the same sample $x$ created by data augmentation techniques. For simplicity of notation, we omit the explicit mention of $z$ network parameters, since it is not utilized by downstream tasks. The BarlowTwins method eliminates the need for explicit negative samples and achieves comparable performance while maintaining computational efficiency. It assumes that negatives are accessible in each mini-batch to estimate correlations among all samples in it.

The network is trained by minimizing an invariance and a redundancy reduction term in the loss function~\cite{zbontar2021barlow}. Here, different augmented views $X_A$ and $X_B$ of the same data samples $X$ are taken from the set of data augmentations $\mathcal{D^*}$. This leads to the loss defined as:
\begin{equation}
\mathcal{L}_c = \mathbb{E}_{X_A, X_B \sim \mathcal{D^*} } \!\! \Big[ \sum_i  (1-\mathcal{C}_{ii})^2 + \lambda \sum_{i} \sum_{j \neq i} {\mathcal{C}_{ij}}^2 \Big],
\label{eq:lossBarlow}
\end{equation}
where $\lambda$ is a positive constant trade-off parameter between both terms, and where $\mathcal{C}$ is the cross-correlation matrix computed between the representations $z$ of all samples $X_A$ and $X_B$ in a mini-batch indexed by $b$:

$
\mathcal{C}_{ij} =  \sum_b z^A_{b,i} z^B_{b,j} / (\sqrt{\sum_b {(z^A_{b,i})}^2} \sqrt{\sum_b {(z^B_{b,j})}^2})
$.
The cross-correlation matrix $\mathcal{C}$ contains values ranging from -1.0 (worst) to 1.0 (best) for the correlation between the projector's outputs: $Z_A = z(g_\phi(X_A))$ and $Z_B = z(g_\phi(X_B))$. The invariance term of the loss function encourages the diagonal elements to have a value of 1. This ensures that the learned embedding is invariant to the applied data augmentations. Meanwhile, the second term (redundancy reduction) maintains the off-diagonal elements close to zero and decorrelates the outputs of non-related images.

\subsection{Continual SSL Problem Definition}

In this work, we consider a CL scenario in which the feature extractor $f_{\theta}$ must learn from a set of task $\{1, \dots, T\}$ from different distributions, where each task $t$ from that set follows the distributions $\mathcal{D}_t$. We would like to find the parameters $\theta$ of the feature extractor $f_{\theta}$ that minimizes the summed loss over all tasks $T$:
\begin{equation}
    \arg\min_{\theta}\sum_{t=1}^T\mathcal{L}_c^t,
\label{eq:cl}
\end{equation}
where $\mathcal{L}_c^t=\mathbb{E}_{X_A, X_B \sim \mathcal{D}^*_t } [\mathcal{L}_c]$ and $\mathcal{L}_c$ is defined as in Eq.~\ref{eq:lossBarlow}. However, finding the right $\theta$ poses the main problem in continual learning, as previous data $D_{1},...,D_{t-1}$ is not available at time $t$ and Eq.~\ref{eq:cl} cannot be minimized directly.

\subsection{Plasticity-Optimized Complementary Networks}

We propose Plasticity-Optimized Complementary Networks (POCON) based on CLS framework. POCON is composed of three stages training (see Fig.~\ref{fig:main_method} for details): \circled{1} learn expert knowledge from current data task, \circled{2} integrate new expert knowledge to the main network, and \circled{3} initialize the new expert from the updated main network. Each stage is explained in the details in the next sections.  

\minisection{\textit{Stage} $\textcircled{1}$: SSL of expert for the current task.}

In this step, we are interested to fully-adapt to the input training data of the current task. Hence, a feature extractor $g_\theta^t$ is used to learn in a self-supervised way (following Eq.~\ref{eq:lossBarlow}) on the data ${D}_t$. Note that unlike previous methods (PFR and CaSSLe), we do not constraint in any way our expert during the training (like imposing regularization to prevent forgetting). We allow the expert network to be fully plastic and optimal for learning representation for the current task.

\minisection{\textit{Stage} \textcircled{2}: Knowledge integration.}
Once we absorb the knowledge of the current task in the expert network at \textit{Stage} \numcircledpict{1}, it needs to be transferred and accumulated to the main feature extractor $f_{\theta}$ without forgetting previous tasks (see Fig.\ref{fig:main_method} \textcircled{2}). To do so, we employ an additional \emph{adaptation projector} $n:\mathcal{Z}\rightarrow \mathcal{W}$, which maps the embedding space from the main network $f_{\theta}$ to the embedding space learned in the expert network $g_{\phi^t}$. Then, to avoid forgetting, we use a \emph{retrospection projector} $m:\mathcal{Z}\rightarrow \mathcal{Z}$ that maps the embedding space learned on the current task back to the embedding space learned on the previous ones. The final loss function for the knowledge integration stage consists of an adaptation and retrospection component:
\begin{multline}
     \mathcal{L}_{INT}^t = \mathbb{E}_{X_A \sim \mathcal{D}^*_t } \Bigg[ \sum_{i=0}^{|X_A|}   \parallel n(g_{\phi^t}\left(x_a\right))-f_{\theta^{t}}\left(x_a\right) \parallel +  \\
         \parallel m(f_{\theta^t}\left(x_a\right))-f_{\theta^{t-1}}\left(x_a\right) \parallel \Bigg] 
     \label{eq:Stage2}
\end{multline}

where both sources from which we integrate our knowledge: previous main network $f_{\theta^{t-1}}$ and the current expert $g_{\phi^t}$ are frozen, and only $f_{\theta^{t}}$ and adaptor networks $n$ and $m$ are being updated by this loss function. The goal is to integrate knowledge using distillation and current task data.

\minisection{\textit{Stage} \textcircled{3}: New expert initialization.}
In order to begin the training on the next task, the expert $g_\phi^{t+1}$ must be prepared to solve the new task with the best prior knowledge for training new task representation efficiently. In this stage, we need to initialize a new expert (see Fig.~\ref{fig:main_method} \textcircled{3}) in the best way. Improper initialization can influence the training epochs in \textit{Stage} \textcircled{1} and make the problem of adaptation for projector $n$ more difficult in \textit{Stage} \textcircled{2}.  We looked into two potential initialization setups based on the similarities between the expert and main backbones. The homogeneous setup, where both the main network and the expert network share the same architecture. This is the default setup in our experiments. In addition, we will also consider the heterogeneous setup, where the expert has another architecture than the main network. This allows, for example, to apply smaller networks when per task data is limited, or computation should be performed on edge devices.

\noindent
\emph{Homogeneous setup (CopyOP):} In order to begin the training of $g_\phi^{t+1}$ with all the accumulated knowledge till $t$ we can just copy the weight of the main network $f_\theta^{t}$ into $g_\phi^{t+1}$. This operation avoids the recency bias of ($g_\phi^{t}$) and provides an excellent initialization point for $g_\phi^{t+1}$ to continue learning. Furthermore, CopyOP makes the problem of the adaptation projector $n$ easier since $g_\phi^{t+1}$ will have a similar representation to $f_\theta^{t}$. The main drawback of CopyOP is that it constrains of POCON to use the same architecture for the main $f_\theta^{t}$ and the expert $g_\phi^{t+1}$ networks.

\noindent
\emph{Heterogeneous setup (D2eOP):} This distillation-based initialization method allows the use of heterogeneous network architectures for $f_\theta^{t}$ and $g_\phi^{t+1}$ in POCON. In order to transfer the knowledge, we propose a projected distillation as in \textit{Stage} \textcircled{2} using a fresh adaptation projector $n$. Despite being more computationally demanding, this way of transferring knowledge offers one big advantage – different architecture for $g_\phi^{t+1}$ allows the use of a smaller backbone network or even very different ones, e.g.~ViT. Using a smaller backbone is useful for low-data regimes (as we will show later) or for devices low in computational power at the time of learning the expert (robotics, edge devices). 

The loss function for D2eOP in \textit{Stage} \textcircled{3} is given as:
\begin{equation}
     \mathcal{L}_{D2eOP}^t = \mathbb{E}_{X_A \sim \mathcal{D}^*_t } \Bigg[ \sum_{i=1}^{|X_A|} \parallel n(g_\phi^{t+1}\left(x_a\right))-f_{\theta^{t}}\left(x_a\right) \parallel \Bigg] 
     \label{eq:Stage3}
\end{equation}

where $n$ is a projector that adapts the embedding space of new expert $g_\phi^{t+1}$ to the previous one from the main network with current task data $D^t$. The difference of D2eOP and CopyOP is presented in the right column of Fig.~\ref{fig:main_method}. 

Other initialization options exists. We discuss them and provide more results in the Appendix.

\subsection{Plasticity-stability trade-off in CURL with SSL}

\begin{figure}
     \centering
     \begin{subfigure}[b]{0.235\textwidth}
         \centering
         \includegraphics[width=\textwidth]{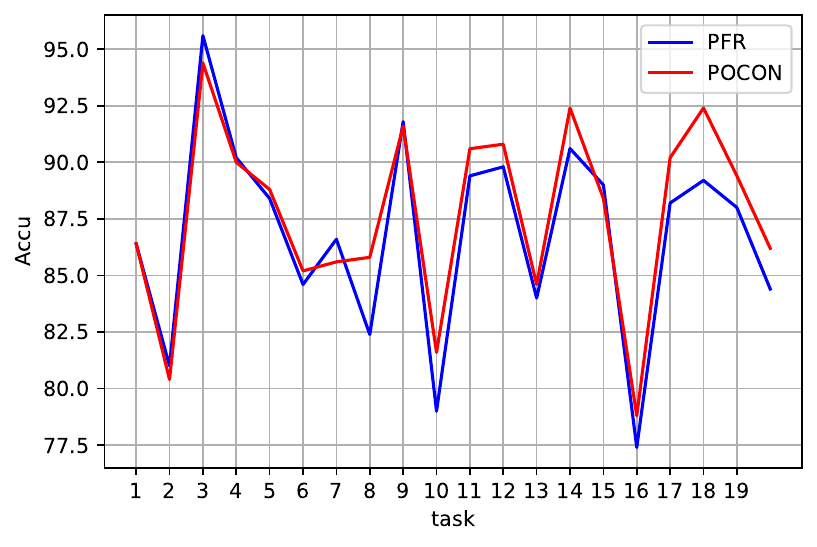}
         \caption{Plasticity}
         
     \end{subfigure}
     \hfill
     \begin{subfigure}[b]{0.235\textwidth}
         \centering
         \includegraphics[width=\textwidth]{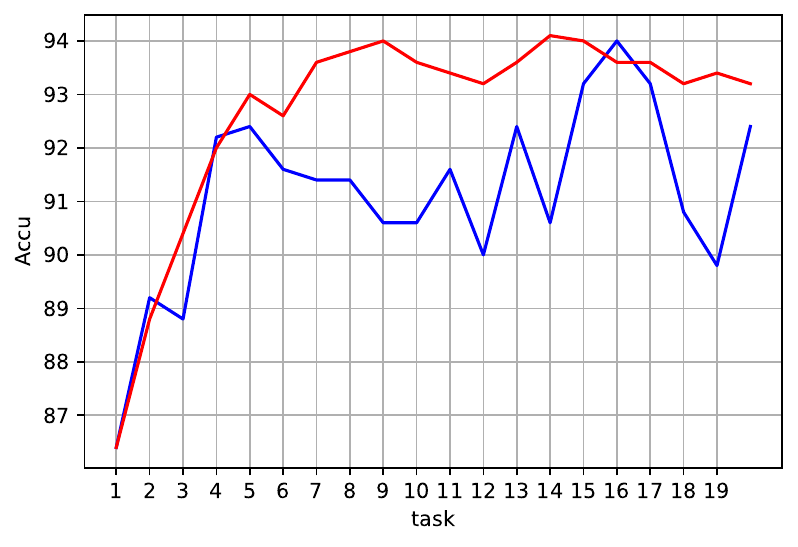}
         \caption{Stability task 1}
         
     \end{subfigure}     
     \begin{subfigure}[b]{0.235\textwidth}
         \centering
         \includegraphics[width=\textwidth]{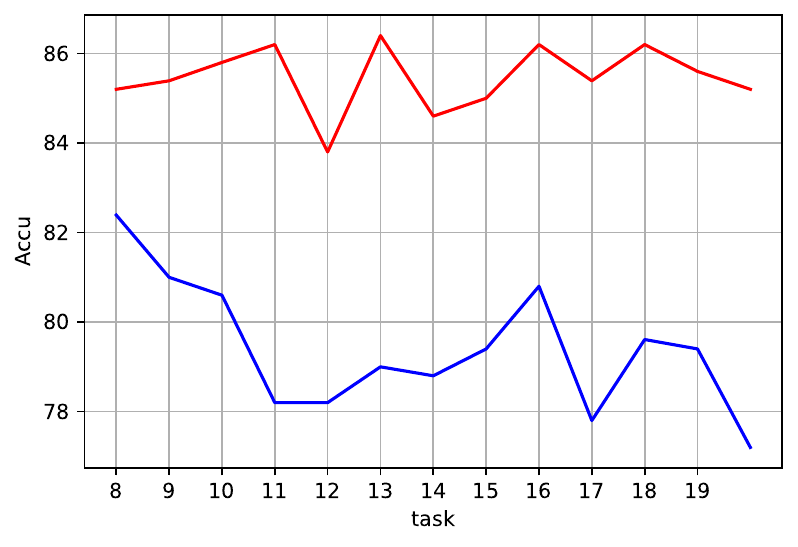}
         \caption{Stability task 8}
         
     \end{subfigure}
     \hfill
     \begin{subfigure}[b]{0.235\textwidth}
         \centering
         \includegraphics[width=\textwidth]{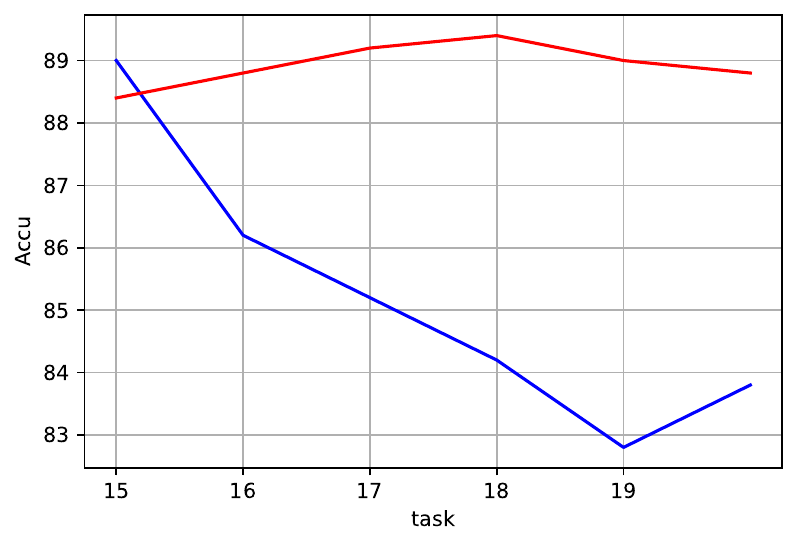}
         \caption{Stability task 15}
         
     \end{subfigure}

        \caption{Plasticity-stability trade-off in POCON and PFR: (a) accuracy of a current task during the continual learning session, (b-d) accuracy of task 1, 8, and 15 in the following incremental steps of learning the representation with POCON and PFR. POCON presents superior stability to PFR at the same time still maintaining better plasticity, with its non-restrictive training of the expert network.}
        \label{fig:stability_plasticity}
\end{figure}

Our work is motivated by two key observations:
\begin{itemize}
    \item Regularized-based CURL models (PFR and CaSSLe) have an implicit handicapped plasticity. During the training, they learn new tasks data using SSL while maintaining the backbone representations close to the previous tasks to avoid forgetting. Therefore, the backbone network cannot fully adapt to current data due to the regularization imposed by the CL method.
    \item As the number of tasks increases, current CURL models lose stability. In this case, the data distribution changes more abruptly and the backbone is pushed to follow an imperfect estimation of the new distribution restricted by the regularization of the old model (see first observation). Hence, these models are not as plastic as finetuning and do not present stable behavior.
    
\end{itemize}

In Fig.~\ref{fig:stability_plasticity} we present the stability-plasticiy tradeoff for twenty task partition of CIFAR100 in PFR and our method (POCON). For the plasticity plot, we evaluate the accuracy at the end of each task $t$ using task $t$ data for training and evaluation. The stability plots were made by training and evaluating in a fixed task partition ($1$, $8$, and $15$ for Fig.~\ref{fig:stability_plasticity}) after the training of each task $t$.

As Fig.~\ref{fig:stability_plasticity} $a)$ shows POCON has a higher accuracy (more plastic) in most of the task during the whole training. As training an expert in stage \circled{1} is not constrained by any regularization. Morover, Fig.~\ref{fig:stability_plasticity} $b)$, $c)$, and $d)$ display how POCON is able to retain previous representations over the whole incremental learning session, where our double distillation (adaptation and retrospection) is able to retain the learned representation correctly.

\section{Experimental Results}\label{sec:Results}

\subsection{Experimental setup}

\minisection{Datasets} We use the following datasets:
\emph{CIFAR-100}~\cite{krizhevsky2009learning},  consists of 100 object classes in 45,000 images for training, 5,000 for validation, and 10,000 for test with 100 classes. All images are 32$\times$32 pixels;
\emph{TinyImageNet} a rescaled subset of 200 ImageNet~\cite{deng2009imagenet} classes used in~\cite{tinyIM} and containing 64$\times$64 pixel images. Each class has 500 training images, 50 validation images, and 50 test images;
\emph{ImageNet100} a subset of one hundred classes from ImageNet~\cite{deng2009imagenet} that contains $130$k images of size 224$\times$224 pixels.

\minisection{Training procedure}
\label{sec:train}
In all experiments, we train ResNet-18~\cite{he2016deep} (or ResNet-9 for the heterogeneity experiment) for expert and main network using SGD with an initial learning rate of $0.01$ and a weight decay of $0.0001$ for $250$ epochs (200 for ImageNet100)in \textit{Stage} \textcircled{1}. For \textit{Stage} \textcircled{2} same optimization procedures as \textit{Stage} \textcircled{1} is followed but for $500$ epochs (400 for ImageNet100). 

The data augmentation used in \textit{Stage} \textcircled{1} is the same as in BarlowTwins~\cite{chen2020SimCLR}). Based on the self-supervised distillation ideas of \cite{sariyildiz2023improving,navaneet2021simreg}, we use a four-layer MLP projector as  adaptation and retrospection projectors following the architecture of~\cite{navaneet2021simreg}.

Downstream classifiers are by default linear and trained with a CE-loss and Adam optimizer with a learning rate $5e$-$2$ on CIFAR-100, and $3$ on TinyImageNet. We use validation data to implement a patience scheme that lowers the learning rate by a factor of $0.3$ and $0.06$ up to three times while training a downstream task classifier. For ImageNet100 we use the same training and evaluation procedure as~\cite{fini2022self}.

\minisection{Baseline methods}
We only compare 
to exemplar-free methods and exclude methods that require replay from our comparison\footnote{Code available at https://github.com/alviur/pocon\_wacv2024}.

\emph{Fine-tuning (\textbf{FT})}: The network is trained sequentially on each task without access to previous data and with no mitigation of catastrophic forgetting.
\emph{\textbf{Joint}}: We perform joint training with fine-tuning on all data which provides an upper bound. Equivalent to having a single-task scenario.

\emph{\textbf{PFR}}~\cite{gomez2022continually} and \emph{\textbf{CaSSLe}}~\cite{fini2022self} with Barlow Twins: We use the code and hyperparameters provided by the authors, in PFR we used $\lambda=25$ for all the experiments.

In continual semi-supervised learning (section~\ref{sec:semisup}), we consider the following methods. 
Regularization-based methods: Learning without Forgetting (LwF)~\cite{li2017learning}, online Elastic Weight Consolidation (oEWC)~\cite{kirkpatrick2017overcoming},
Replay-based methods: Experience Replay (ER)~\cite{rolnick2019experience}, iCaRL~\cite{rebuffi2017icarl} and GDumbciteprabhu2020gdumb, and
Continual semi-supervised learning methods: CCIC~\cite{boschini2022continual}, PAWS~\cite{assran2021semi} and NNCSL~\cite{kang2022soft}.

\subsection{Continual representation learning}

\begin{table}[t]
\centering
\begin{minipage}{0.46\textwidth}

\centering
\resizebox{\linewidth}{!}{

\begin{tabular}{l|lllll}
\toprule
\multicolumn{6}{c}{CIFAR-100 (32x32)}                                    \\
\midrule
\textbf{Method}  & \textbf{4 tasks} & \textbf{10 tasks} & \textbf{20 tasks} & 5\textbf{0 tasks} & \textbf{100 tasks} \\
\midrule
FT  & 54.8  & 50.94  & 44.95  & 38.0   & 27.0         \\
CaSSLe & 59.80 &52.5 &49.6 & 45.3 & 42.10         \\
PFR  & 59.70  & 54.33  & 44.80  & 46.5  & 43.30         \\

POCON & \textbf{63.7}  & \textbf{60.5}  & \textbf{56.8}  & \textbf{48.9} & \textbf{48.94}         \\
\midrule
Joint   & \multicolumn{5}{c}{65.4}        \\
\bottomrule
\end{tabular}
}

\end{minipage}
\hspace{4pt}
\begin{minipage}{0.47\textwidth}

 \centering

\vspace{5pt}
\resizebox{\linewidth}{!}{
\label{tab:tiny_main}
\begin{tabular}{l|lllll}
\toprule
\multicolumn{6}{c}{TinyImagenet (64x64)}                                    \\
\midrule
\textbf{Method}    &  \textbf{4 tasks}  &  \textbf{10 tasks} & \textbf{20 tasks} & \textbf{50 tasks} & \textbf{100 tasks}\\
\midrule
FT  &41.95      & 36.55        &32.29     & 22.34   & 2.80 \\
\midrule
CaSSLe &\textbf{46.37}     & \textbf{41.53}   & 38.18   & 28.08& 25.38        \\
PFR & 42.23& 39.20 & 31.22& 25.87 & 21.20  \\
POCON &40.97& 41.06  &\textbf{41.14}& \textbf{37.20} & \textbf{30.24} \\
\midrule
Joint     & \multicolumn{5}{c}{50.18}                            \\
\midrule
\multicolumn{6}{c}{ImageNet100 (224x224)}                                    \\
\midrule
\textbf{Method}    &  \textbf{5 tasks}  &  \textbf{10 tasks} & \textbf{20 tasks} & \textbf{50 tasks} & \textbf{100 tasks}\\
\midrule
FT  &56.10      & 48.13        &42.73     & 39.64 & 21.03    \\
\midrule
CaSSLe &\textbf{67.56} & 59.78& 53.92   & 46.64& 36.44        \\
PFR  & 66.12& 60.46 & 54.84& 42.18 & 38.34  \\

POCON  & 66.30 & \textbf{61.36}  & \textbf{59.32}   & \textbf{53.50} &  \textbf{45.40}  \\
\midrule
Joint     & \multicolumn{5}{c}{71.06}                            \\
\bottomrule

\end{tabular}

}

\end{minipage}
\caption{Accuracy of a linear evaluation on equal split various datasets and different number of tasks with ResNet-18. POCON present better result than other regularization-based methods and maintain high accuracy even with increasing number of tasks.}
\label{tab:cifar100_main}

\end{table}

In this experiment, we evaluate all methods in the continual representation learning setting, where each task consist of a distinct set of classes from a single dataset. Splits are prepared similarly to the class incremental learning setting, but without access to labels. Specifically, we split datasets into four, ten, twenty, fifty, and one hundred equal tasks as done in~\cite{rebuffi2017icarl}. In each task, we perform SSL (\textit{Stage} \textcircled{1}), knowledge integration (\textit{Stage} \textcircled{2}), and a new expert initialization (\textit{Stage} \textcircled{3}). In the evaluation phase, we train a linear classifier using the learned representation of the main network encoder. Please note that the expert network is never used for evaluation (unless specified). We use all available test data to obtain the overall task-agnostic performance evaluation\footnote{We use \emph{task-agnostic} evaluation in this paper to refer to the class-incremental learning evaluation~\cite{masana2022class} as in CaSSLe and PFR methods}. In all our tables we report the accuracy in the last task.

\noindent
\textbf{Homogeneous setup.}
Table~\ref{tab:cifar100_main} presents the results for all the methods on three commonly used dataset - CIFAR100, TinyImageNet, and ImageNet100. 
For CIFAR100 the upper bound of Joint training for a single task is $65.4\%$. Note, that the gap between FT and \emph{Joint} is getting bigger with more tasks, from $10.6\%$ for four task up to $38.4\%$ for extreme case of one hundred tasks. POCON is significantly better than CaSSLe and PFR for different number of tasks. In four tasks setting, POCON is only $1.7\%$ points lower than Joint, while second next, CaSSLe reaches $3.3\%$ pints lower accuracy. With the increasing number of tasks, the CaSSLe performance drops faster than PFR, while POCON maintains superior results against others and presents the lowest decrease in performance. 

The results for TinyImagenet and ImageNet100 follow the same procedure as for CIFAR-100, but here we have larger images (64x64 and 224x224) and more classes for TinyImagenet (200). In this case, POCON outperforms other  methods whenever the number of tasks is higher, 20 for TinyImangeNet and 10 for ImageNet100. For a few tasks scenario, POCON will be close to other CL methods due to high data availability. The accuracy gap between POCON and other methods increases with the number of tasks.

\noindent
\textbf{Heterogeneous setup.}
An expert in POCON can use a different network architecture than the main network. That opens the possibility of using a smaller network for the expert whenever this can be beneficial, e.g.,~tasks are small with not enough data to train a large  ResNet-18 network, or the device where the expert network is being trained is not powerful enough (robot, edge). We investigated heterogeneous architecture use in POCON with a smaller network, ResNet-9 which has $6.5$M number of parameters instead of 11M in ResNet-18. The results are presented in Table~\ref{tab:cifar100_res}. 
The different combinations of POCON  are presented for using smaller network in expert only, or for both, expert and main networks. 
With increasing number of tasks it is more beneficial to use smaller expert (20 tasks). And having less data per task (50 and 100 tasks) we see improved results when as well the main network is smaller, we gain $4.6\%$ changing from ResNet-18 to ResNet-9 for one hundred tasks.

\begin{table}[tb]
\centering
\centering
\resizebox{\linewidth}{!}{
\begin{tabular}{ll|lllll}

\toprule
\textbf{Method}  &   \textbf{Arch}  & \textbf{4 tasks} & \textbf{10 tasks} & \textbf{20 tasks} & 5\textbf{0 tasks} & \textbf{100 tasks} \\
\midrule
    & 58.95  & 55.4  &50.77 & 49.18  & 40.78  \\
POCON  & R18-R18 & \textbf{63.7}  & \textbf{60.5}  & 56.8  & 48.9 & 48.94         \\
POCON  & R18-R9  &  62.05 & 60.5 & \textbf{57.48} & 49.7 & 47.94         \\
POCON  & R9-R9 & 58.34 & 58.32  &  56.07 & \textbf{53.3}  & \textbf{51.22}         \\
\bottomrule
\end{tabular}
}
\caption{Accuracy of linear evaluation with heterogeneous POCON architecture using ResNet-18 (R18) and ResNet-9 (R9) on split CIFAR-100. With increasing number of tasks and lower data POCON benefits from a smaller ResNet-9 network, first, only for the expert (20 tasks) and later (50 tasks) for both networks, expert and main. }
\label{tab:cifar100_res}
\end{table}

\begin{table}
\centering
\resizebox{\linewidth}{!}{
\begin{tabular}{c|c|ccc}
\toprule
\multirow{2}{*}{Method}  &  \multirow{2}{*}{exemplar} &     \multicolumn{3}{c}{CIFAR-100}\\
                 &          & 0.8\%       & 5.0\%          & 25.0\%         \\
\hline

\hline
\multicolumn{5}{c}{\textbf{\textit{Exemplar-based methods}}}\\
\hline
ER~\cite{rolnick2019experience} & \cmark   & 8.2 $\pm$ 0.1 & 13.7 $\pm$ 0.6& 17.1 $\pm$ 0.7   \\
iCaRL~\cite{rebuffi2017icarl} & \cmark   & 3.6 $\pm$ 0.1 & 11.3 $\pm$ 0.3   & 27.6 $\pm$ 0.4   \\
GDumb~\cite{prabhu2020gdumb}& \cmark   & 8.6 $\pm$ 0.1    & 9.9 $\pm$ 0.4  & 10.1 $\pm$ 0.4   \\
\hline
CCIC~\cite{boschini2022continual} (500)       & \cmark      & 11.5 $\pm$ 0.7   & 19.5 $\pm$ 0.2   & 20.3 $\pm$ 0.3   \\
PAWS~\cite{assran2021semi} (500)       & \cmark      & 16.1 $\pm$ 0.4   & 21.2 $\pm$ 0.4   & 19.2 $\pm$ 0.4   \\
NNCSL~\cite{kang2022soft} (500)      & \cmark      & \textbf{27.4} $\pm$ 0.5   & \textbf{31.4} $\pm$ 0.4   & \textbf{35.3} $\pm$ 0.3   \\
\hline
\multicolumn{5}{c}{\textbf{\textit{Exemplar-free methods}}}\\
\hline
\hline
Fine-tuning      & \xmark       & 1.8 $\pm$ 0.2    & 5.0 $\pm$ 0.3    & 7.8 $\pm$ 0.1    \\
LwF~\cite{li2017learning} & \xmark  & 1.6 $\pm$ 0.1    & 4.5 $\pm$ 0.1    & 8.0 $\pm$ 0.1    \\
oEWC~\cite{kirkpatrick2017overcoming} & \xmark  & 1.4 $\pm$ 0.1    & 4.7 $\pm$ 0.1    & 7.8 $\pm$ 0.4    \\

\hline
Prototypes        & \xmark        & 19.2  $\pm$  0.9 & 23.5 $\pm$  0.4  & 24.1  $\pm$  0.1 \\
Prototypes\textit{+SDC}         & \xmark        &  \textbf{22.7} $\pm$  0.6 & \textbf{27.6}  $\pm$  0.4 & \textbf{28.5}  $\pm$  0.1 \\

\bottomrule
\end{tabular}
}
\caption{
Semi-supervised continual learning comparison on CIFAR100 dataset. The number between brackets indicates the size of the memory buffer. 
We highlight the best method in each group with \textbf{bold} fonts.
}
\label{tab:semiSup}
\end{table}

\begin{figure*}
     \centering
     \begin{subfigure}[b]{0.32\textwidth}
         \centering
         \includegraphics[width=\textwidth]{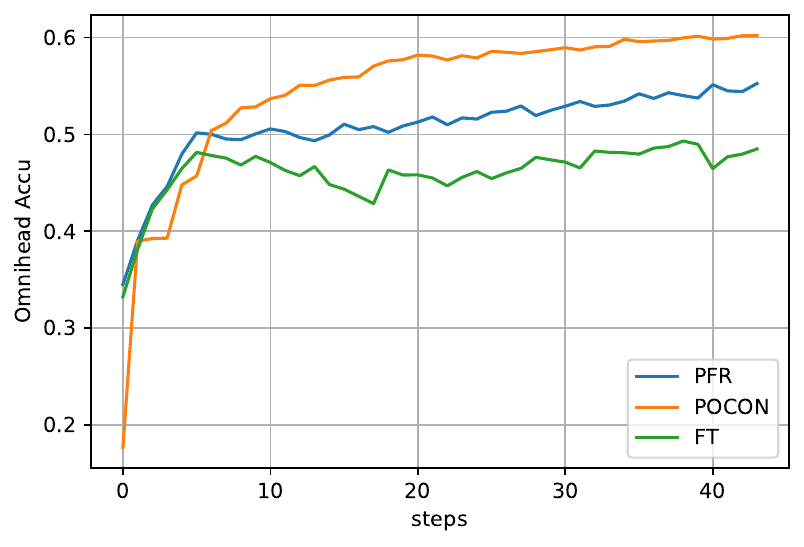}
         \caption{10 blurred data partitions}
         \label{fig:10 blurred}
     \end{subfigure}     
     \begin{subfigure}[b]{0.32\textwidth}
         \centering
         \includegraphics[width=\textwidth]{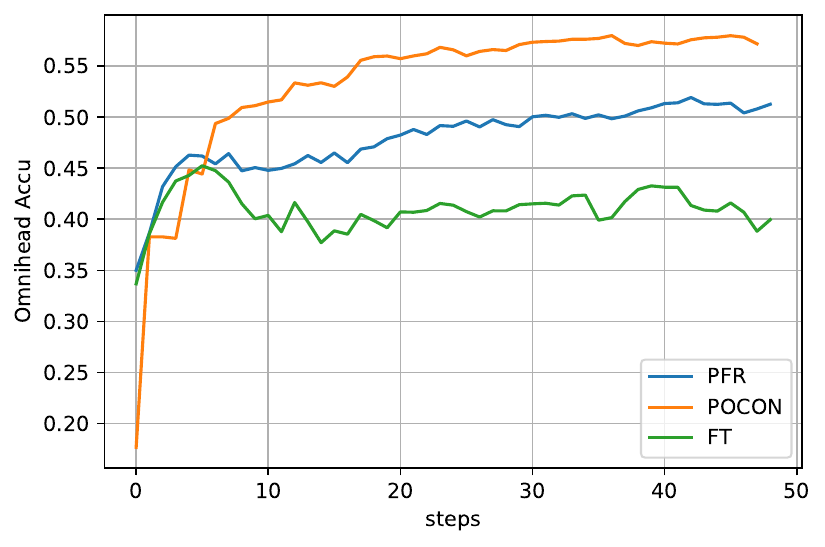}
         \caption{20 blurred data partitions}
         \label{fig20 blurred}
     \end{subfigure}     
     \begin{subfigure}[b]{0.32\textwidth}
         \centering
         \includegraphics[width=\textwidth]{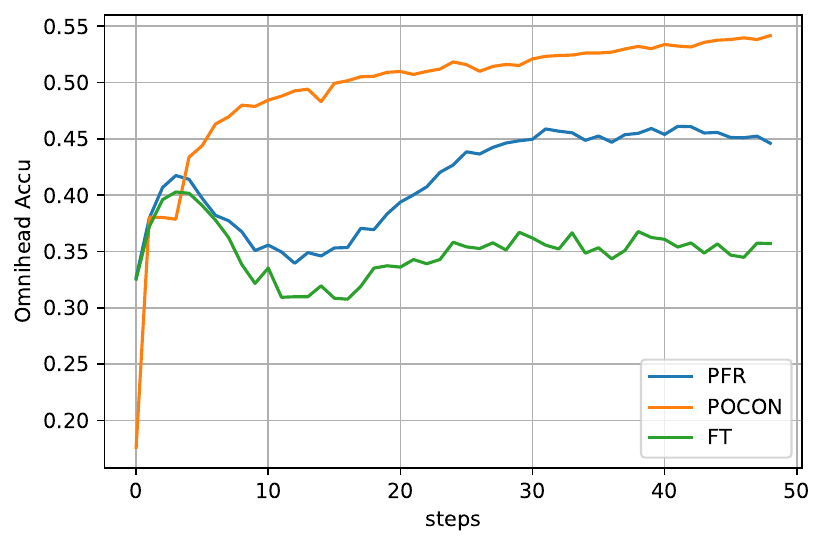}
         \caption{50 blurred data partitions}
         \label{fig:50 blurred}
     \end{subfigure}

        \caption{Task-task setting for CIFAR-100 with different length of incremental learning sequence. Plots present linear evaluation of learned representation for the blurred 10, 20, and 30 data partitions respectively (like in~\cite{10.1162/neco_a_01430, li2022energy}). POCON presents significantly better results with stable learning accuracy curves.}
  \label{fig:blurred}
\end{figure*}

\subsection{Task-free setting}

To show the ability of POCON to handle varying data streams, we test it in the task-free setting. In this setting, there is no explicit boundary between tasks, the data from one task changes smoothly to the data from the next~\cite{10.1162/neco_a_01430,li2022energy}. This prevents methods from having a fixed point where the network can change and prepare for the new task; the adaptation is ongoing. For instance, when we receive data $D_t$ there is a mix with the data $D_{t-1}$. At some point, we only get data from $D_t$, but later on, we will get a mix with $D_{t+1}$. For this experiment, we employ the data partition of~\cite{10.1162/neco_a_01430} for the CIFAR100 dataset with beta equal to $4$ (please see the Appendix for more details).

Since there is no clear boundary, we cannot perform distillation without losing data from the stream of mini-batches. In this setting, POCON performs \textit{Stage} \textcircled{2} in parallel with \textit{Stage} \textcircled{1}. In order to do so, a frozen copy of the expert $g_\phi$ is used for the \textit{Stage} \textcircled{2} while a new expert is learning on the current data. After $s$ steps, a copy of the expert $g_\phi$ is passed to perform a distillation for $ds$ steps. Note that we do not store any data; distillation for \textit{Stage} \textcircled{2} and SSL in \textit{Stage} \textcircled{1} is always performed using the current mini-batch data. \textit{Stage} \textcircled{3} is omitted, as the expert network is not changed and initialized, as there is no task switch.  

To compare to other method, we adapted PFR to work in the task-free setting similarly. In this case, the feature extractor of past data is updated after $s$ steps (copy of the current feature extractor), and the regularization is performed during the whole training as in normal PFR. We also present results with a simple fine-tuning (FT).

Fig.~\ref{fig:blurred} presents the results of linear evaluation of the learned representation in continual learning for the ten, twenty, and fifty data partitions settings. 
Only at the first steps at the beginning, for 10 and 20 tasks, PFR and FT accuracy is above POCON. However, that changes significantly in the following steps in favor of POCON, which continues improving learned representations. Unlike PFR and FT, POCON has a more stable learning curve for all task splits. The improvements over PFR and FT are better for more tasks, since the regularization-based method hurts plasticity while learning new tasks with PFR.

\subsection{Continual semi-supervised learning}\label{sec:semisup}

We propose a simple extension of POCON to the semi-supervised setting, where a small percentage of the data is labeled. We train the method in the same fashion as the unsupervised case, and we initially ignore  available labels. After updating $f_\theta^{t}$ until convergence with POCON, we initialize the prototypes of each class as the average of the labeled samples. Then we assign all unlabeled data to the nearest prototype center, and we again compute the average of all samples assigned to each prototype. We perform the same procedure after each of the tasks. We call this method \emph{Prototypes}. We also show for \emph{Prototypes+SDC} where we use Semantic Drift Compensation~\cite{yu2020semantic} to prevent forgetting - we estimate and compensate for learned class prototypes drift.

In Table~\ref{tab:semiSup} we present the results for continual semi-supervised learning under three levels of supervision, namely with only 0.8\%, 5.0\% and 25.0\% of samples labeled (following the settings of NNCSL~\cite{kang2022soft} and CCIC~\cite{boschini2022continual}).
Simply applying prototypes with POCON achieves much better results than the other exemplar-free methods (LwF~\cite{li2017learning} and oEWC~\cite{kirkpatrick2017overcoming}). The method outperforms the continual learning methods which only exploit the labeled data (such as ER, iCaRL, and GDumb). Furthermore, it also outperforms the semi-supervised methods CCIC and PAWS. Note that our method, without storing any exemplars, is only outperformed by the recent NNCSL method which requires exemplars and is dedicated to the semi-supervised learning setting.

\section{Conclusions and Future directions}

We proposed a method for exemplar-free continual unsupervised representation learning called \textit{Plasticity-Optimized COmplementary Networks}. POCON trains an expert network that performs optimally on the current data without any regularization (optimizing plasticity). The knowledge is transferred to the main network through distillation while retaining similarity with the old main network to avoid forgetting. Experiments of CIFAR100, TinyImagenet, and ImageNet100 show that our approach outperforms other methods for SSL exemplar-free CL learning (PFR and CaSSLe), and it is especially good for many tasks scenarios.

The POCON method presents several opportunities for improvement. One promising direction is to extend the method to the scenario where multiple experts can be trained in parallel on different tasks, similar to the clients in federated learning. Secondly, heterogeneity of the fast and slow learner can be better investigated – how we can benefit from having different architectures (even mixed ones with transformer-based network). 

\paragraph{Limitations.}

Although we show how to squeeze (and retain) the knowledge learned by the SSL method, when the number of samples per tasks is too low, the knowledge transfer (stages \textcircled{2} and \textcircled{3}) and the expert training (stage \textcircled{1}) degrades. Note that in this extreme scenario, we are going towards online continual learning.

\paragraph{Impact Statement.} Continual learning systems do not  require data to be stored. As such, they can contribute to data privacy, and reduce vulnerabilities related to data storage. As with all machine learning algorithms, special care should be taken to address biases present in the data (and the data collection process). Continual learning could exacerbate biases in the data because of the task-recency bias, which refers to the problem that continual learning algorithms tend to be biased towards the last data fed to the algorithm. 

\paragraph*{Acknowledgments.} We acknowledge the support from the Spanish Government funding for projects PID2022-143257NB-I00, TED2021-132513B-I00 funded by MCIN/AEI/10.13039/501100011033 and by FSE+ and the European Union NextGenerationEU/PRTR, and the CERCA Programme of Generalitat de Catalunya. Bartłomiej Twardowski acknowledges the grant RYC2021-032765-I.

\small
\bibliographystyle{plain}
\bibliography{egbib}

\begin{thebibliography}{10}

\bibitem{aljundi2018memory}
Rahaf Aljundi, Francesca Babiloni, Mohamed Elhoseiny, Marcus Rohrbach, and
  Tinne Tuytelaars.
\newblock Memory aware synapses: Learning what (not) to forget.
\newblock In {\em European Conference on Computer Vision}, 2018.

\bibitem{arani2022learning}
Elahe Arani, Fahad Sarfraz, and Bahram Zonooz.
\newblock Learning fast, learning slow: A general continual learning method
  based on complementary learning system.
\newblock In {\em International Conference on Learning Representations}, 2022.

\bibitem{assran2021semi}
Mahmoud Assran, Mathilde Caron, Ishan Misra, Piotr Bojanowski, Armand Joulin,
  Nicolas Ballas, and Michael Rabbat.
\newblock Semi-supervised learning of visual features by non-parametrically
  predicting view assignments with support samples.
\newblock In {\em Proceedings of the IEEE/CVF International Conference on
  Computer Vision}, pages 8443--8452, 2021.

\bibitem{bang2021rainbow}
Jihwan Bang, Heesu Kim, YoungJoon Yoo, Jung-Woo Ha, and Jonghyun Choi.
\newblock Rainbow memory: Continual learning with a memory of diverse samples.
\newblock In {\em Proceedings of the IEEE/CVF Conference on Computer Vision and
  Pattern Recognition}, pages 8218--8227, 2021.

\bibitem{bardes2022vicreg}
Adrien Bardes, Jean Ponce, and Yann LeCun.
\newblock {VICR}eg: Variance-invariance-covariance regularization for
  self-supervised learning.
\newblock In {\em International Conference on Learning Representations}, 2022.

\bibitem{boschini2022continual}
Matteo Boschini, Pietro Buzzega, Lorenzo Bonicelli, Angelo Porrello, and Simone
  Calderara.
\newblock Continual semi-supervised learning through contrastive interpolation
  consistency.
\newblock {\em Pattern Recognition Letters}, 162:9--14, 2022.

\bibitem{caron2020SwAV}
Mathilde Caron, Ishan Misra, Julien Mairal, Priya Goyal, Piotr Bojanowski, and
  Armand Joulin.
\newblock Unsupervised learning of visual features by contrasting cluster
  assignments.
\newblock {\em arXiv preprint arXiv:2006.09882}, 2020.

\bibitem{chaudhry2018riemannian}
Arslan Chaudhry, Puneet~K Dokania, Thalaiyasingam Ajanthan, and Philip~HS Torr.
\newblock Riemannian walk for incremental learning:~understanding forgetting
  and intransigence.
\newblock In {\em European Conference on Computer Vision}, 2018.

\bibitem{chaudhry2018efficient}
Arslan Chaudhry, Marc’Aurelio Ranzato, Marcus Rohrbach, and Mohamed
  Elhoseiny.
\newblock Efficient lifelong learning with a-gem.
\newblock In {\em International Conference on Learning Representations}, 2019.

\bibitem{chaudhry2019tiny}
Arslan Chaudhry, Marcus Rohrbach, Mohamed Elhoseiny, Thalaiyasingam Ajanthan,
  Puneet~K Dokania, Philip~HS Torr, and Marc'Aurelio Ranzato.
\newblock On tiny episodic memories in continual learning.
\newblock {\em arXiv preprint arXiv:1902.10486}, 2019.

\bibitem{chen2020SimCLR}
Ting Chen, Simon Kornblith, Mohammad Norouzi, and Geoffrey Hinton.
\newblock A simple framework for contrastive learning of visual
  representations.
\newblock In {\em International conference on machine learning}, pages
  1597--1607. PMLR, 2020.

\bibitem{chen2021SimSiam}
Xinlei Chen and Kaiming He.
\newblock Exploring simple siamese representation learning.
\newblock In {\em Proceedings of the IEEE/CVF Conference on Computer Vision and
  Pattern Recognition}, pages 15750--15758, 2021.

\bibitem{cheraghian2021semantic}
Ali Cheraghian, Shafin Rahman, Pengfei Fang, Soumava~Kumar Roy, Lars Petersson,
  and Mehrtash Harandi.
\newblock Semantic-aware knowledge distillation for few-shot class-incremental
  learning.
\newblock In {\em Proceedings of the IEEE/CVF Conference on Computer Vision and
  Pattern Recognition}, pages 2534--2543, 2021.

\bibitem{delange2021continual}
Matthias Delange, Rahaf Aljundi, Marc Masana, Sarah Parisot, Xu~Jia, Ales
  Leonardis, Greg Slabaugh, and Tinne Tuytelaars.
\newblock A continual learning survey: Defying forgetting in classification
  tasks.
\newblock {\em IEEE Transactions on Pattern Analysis and Machine Intelligence},
  2021.

\bibitem{deng2009imagenet}
Jia Deng, Wei Dong, Richard Socher, Li-Jia Li, Kai Li, and Li~Fei-Fei.
\newblock Imagenet: A large-scale hierarchical image database.
\newblock In {\em Conference on Computer Vision and Pattern Recognition}, 2009.

\bibitem{dhar2019learning}
Prithviraj Dhar, Rajat~Vikram Singh, Kuan-Chuan Peng, Ziyan Wu, and Rama
  Chellappa.
\newblock Learning without memorizing.
\newblock In {\em Conference on Computer Vision and Pattern Recognition}, 2019.

\bibitem{doersch2015_pretext_patches}
Carl Doersch, Abhinav Gupta, and Alexei~A Efros.
\newblock Unsupervised visual representation learning by context prediction.
\newblock In {\em Proceedings of the IEEE international conference on computer
  vision}, pages 1422--1430, 2015.

\bibitem{douillard2020podnet}
Arthur Douillard, Matthieu Cord, Charles Ollion, Thomas Robert, and Eduardo
  Valle.
\newblock Podnet: Pooled outputs distillation for small-tasks incremental
  learning.
\newblock In {\em Computer Vision--ECCV 2020: 16th European Conference,
  Glasgow, UK, August 23--28, 2020, Proceedings, Part XX 16}, pages 86--102.
  Springer, 2020.

\bibitem{fini2022self}
Enrico Fini, Victor G~Turrisi Da~Costa, Xavier Alameda-Pineda, Elisa Ricci,
  Karteek Alahari, and Julien Mairal.
\newblock Self-supervised models are continual learners.
\newblock In {\em Proceedings of the IEEE/CVF Conference on Computer Vision and
  Pattern Recognition}, pages 9621--9630, 2022.

\bibitem{gallardo2021self}
Jhair Gallardo, Tyler~L Hayes, and Christopher Kanan.
\newblock Self-supervised training enhances online continual learning.
\newblock {\em arXiv preprint arXiv:2103.14010}, 2021.

\bibitem{gidaris2018_pretext_rotation}
Spyros Gidaris, Praveer Singh, and Nikos Komodakis.
\newblock Unsupervised representation learning by predicting image rotations.
\newblock {\em arXiv preprint arXiv:1803.07728}, 2018.

\bibitem{gomez2022continually}
Alex Gomez-Villa, { Twardowski, Bartlomiej}, { Yu, Lu}, Andrew~D Bagdanov, and
  Joost van~de Weijer.
\newblock Continually learning self-supervised representations with projected
  functional regularization.
\newblock In {\em Proceedings of the IEEE/CVF Conference on Computer Vision and
  Pattern Recognition}, pages 3867--3877, 2022.

\bibitem{NEURIPS2020BYOL}
Jean-Bastien Grill, Florian Strub, Florent Altch\'{e}, Corentin Tallec, Pierre
  Richemond, Elena Buchatskaya, Carl Doersch, Bernardo Avila~Pires, Zhaohan
  Guo, Mohammad Gheshlaghi~Azar, Bilal Piot, koray kavukcuoglu, Remi Munos, and
  Michal Valko.
\newblock Bootstrap your own latent - a new approach to self-supervised
  learning.
\newblock In H.~Larochelle, M.~Ranzato, R.~Hadsell, M.~F. Balcan, and H.~Lin,
  editors, {\em Advances in Neural Information Processing Systems}, volume~33,
  pages 21271--21284. Curran Associates, Inc., 2020.

\bibitem{he2016deep}
Kaiming He, Xiangyu Zhang, Shaoqing Ren, and Jian Sun.
\newblock Deep residual learning for image recognition.
\newblock In {\em Conference on Computer Vision and Pattern Recognition}, 2016.

\bibitem{hou2019learning}
Saihui Hou, Xinyu Pan, Chen~Change Loy, Zilei Wang, and Dahua Lin.
\newblock Learning a unified classifier incrementally via rebalancing.
\newblock In {\em International Conference on Computer Vision}, 2019.

\bibitem{hu2021distilling}
Xinting Hu, Kaihua Tang, Chunyan Miao, Xian-Sheng Hua, and Hanwang Zhang.
\newblock Distilling causal effect of data in class-incremental learning.
\newblock In {\em Proceedings of the IEEE/CVF Conference on Computer Vision and
  Pattern Recognition}, pages 3957--3966, 2021.

\bibitem{kang2022soft}
Zhiqi Kang, Enrico Fini, Moin Nabi, Elisa Ricci, and Karteek Alahari.
\newblock A soft nearest-neighbor framework for continual semi-supervised
  learning.
\newblock {\em arXiv preprint arXiv:2212.05102}, 2022.

\bibitem{kemker2017fearnet}
Ronald Kemker and Christopher Kanan.
\newblock Fearnet: Brain-inspired model for incremental learning.
\newblock {\em ICLR2018}, 2018.

\bibitem{kirkpatrick2017overcoming}
James Kirkpatrick, Razvan Pascanu, Neil Rabinowitz, Joel Veness, Guillaume
  Desjardins, Andrei~A Rusu, Kieran Milan, John Quan, Tiago Ramalho, Agnieszka
  Grabska-Barwinska, et~al.
\newblock Overcoming catastrophic forgetting in neural networks.
\newblock {\em National Academy of Sciences}, 2017.

\bibitem{krizhevsky2009learning}
Alex Krizhevsky, Geoffrey Hinton, et~al.
\newblock Learning multiple layers of features from tiny images.
\newblock 2009.

\bibitem{kumaran2016learning}
Dharshan Kumaran, Demis Hassabis, and James~L McClelland.
\newblock What learning systems do intelligent agents need? complementary
  learning systems theory updated.
\newblock {\em Trends in cognitive sciences}, 20(7):512--534, 2016.

\bibitem{li2022energy}
Shuang Li, Yilun Du, Gido van~de Ven, and Igor Mordatch.
\newblock Energy-based models for continual learning.
\newblock In {\em Conference on Lifelong Learning Agents}, pages 1--22. PMLR,
  2022.

\bibitem{li2017learning}
Zhizhong Li and Derek Hoiem.
\newblock Learning without forgetting.
\newblock {\em IEEE Transactions on Pattern Analysis and Machine Intelligence},
  2017.

\bibitem{liu2018rotate}
Xialei Liu, Marc Masana, Luis Herranz, Joost Van~de Weijer, Antonio~M Lopez,
  and Andrew~D Bagdanov.
\newblock Rotate your networks: Better weight consolidation and less
  catastrophic forgetting.
\newblock In {\em International Conference on Pattern Recognition}, 2018.

\bibitem{madaan2022representational}
Divyam Madaan, Jaehong Yoon, Yuanchun Li, Yunxin Liu, and Sung~Ju Hwang.
\newblock Representational continuity for unsupervised continual learning.
\newblock In {\em International Conference on Learning Representations}, 2022.

\bibitem{mallya2018piggyback}
Arun Mallya, Dillon Davis, and Svetlana Lazebnik.
\newblock Piggyback: Adapting a single network to multiple tasks by learning to
  mask weights.
\newblock In {\em European Conference on Computer Vision}, 2018.

\bibitem{mallya2018packnet}
Arun Mallya and Svetlana Lazebnik.
\newblock Packnet: Adding multiple tasks to a single network by iterative
  pruning.
\newblock In {\em Conference on Computer Vision and Pattern Recognition}, 2018.

\bibitem{masana2022class}
Marc Masana, Xialei Liu, Bart{\l}omiej Twardowski, Mikel Menta, Andrew~D
  Bagdanov, and Joost Van De~Weijer.
\newblock Class-incremental learning: survey and performance evaluation on
  image classification.
\newblock {\em IEEE Transactions on Pattern Analysis and Machine Intelligence},
  45(5):5513--5533, 2022.

\bibitem{masana2020ternary}
Marc Masana, Tinne Tuytelaars, and Joost van~de Weijer.
\newblock Ternary feature masks: continual learning without any forgetting.
\newblock {\em arXiv}, 2020.

\bibitem{mcclelland1995there}
James~L McClelland, Bruce~L McNaughton, and Randall~C O'Reilly.
\newblock Why there are complementary learning systems in the hippocampus and
  neocortex: insights from the successes and failures of connectionist models
  of learning and memory.
\newblock {\em Psychological review}, 102(3):419, 1995.

\bibitem{mccloskey1989catastrophic}
Michael McCloskey and Neal~J Cohen.
\newblock Catastrophic interference in connectionist networks: The sequential
  learning problem.
\newblock In {\em Psychology of learning and motivation}, volume~24, pages
  109--165. Elsevier, 1989.

\bibitem{navaneet2021simreg}
K~L Navaneet, Soroush~Abbasi Koohpayegani, Ajinkya Tejankar, and Hamed
  Pirsiavash.
\newblock Simreg: Regression as a simple yet effective tool for self-supervised
  knowledge distillation.
\newblock In {\em British Machine Vision Conference (BMVC)}, 2021.

\bibitem{noroozi2016_pretext_puzzle}
Mehdi Noroozi and Paolo Favaro.
\newblock Unsupervised learning of visual representations by solving jigsaw
  puzzles.
\newblock In {\em European conference on computer vision}, pages 69--84.
  Springer, 2016.

\bibitem{o2002hippocampal}
Randall~C O'Reilly and Kenneth~A Norman.
\newblock Hippocampal and neocortical contributions to memory: Advances in the
  complementary learning systems framework.
\newblock {\em Trends in cognitive sciences}, 6(12):505--510, 2002.

\bibitem{parisi2019continual}
German~I Parisi, Ronald Kemker, Jose~L Part, Christopher Kanan, and Stefan
  Wermter.
\newblock Continual lifelong learning with neural networks: A review.
\newblock {\em Neural networks}, 113:54--71, 2019.

\bibitem{parisi2018lifelong}
German~I Parisi, Jun Tani, Cornelius Weber, and Stefan Wermter.
\newblock Lifelong learning of spatiotemporal representations with dual-memory
  recurrent self-organization.
\newblock {\em Frontiers in neurorobotics}, page~78, 2018.

\bibitem{pham2021dualnet}
Quang Pham, Chenghao Liu, and Steven Hoi.
\newblock Dualnet: Continual learning, fast and slow.
\newblock {\em Advances in Neural Information Processing Systems},
  34:16131--16144, 2021.

\bibitem{prabhu2020gdumb}
Ameya Prabhu, Philip~HS Torr, and Puneet~K Dokania.
\newblock Gdumb: A simple approach that questions our progress in continual
  learning.
\newblock In {\em Computer Vision--ECCV 2020: 16th European Conference,
  Glasgow, UK, August 23--28, 2020, Proceedings, Part II 16}, pages 524--540.
  Springer, 2020.

\bibitem{rebuffi2017icarl}
Sylvestre-Alvise Rebuffi, Alexander Kolesnikov, Georg Sperl, and Christoph~H
  Lampert.
\newblock icarl: Incremental classifier and representation learning.
\newblock In {\em Conference on Computer Vision and Pattern Recognition}, 2017.

\bibitem{rolnick2019experience}
David Rolnick, Arun Ahuja, Jonathan Schwarz, Timothy Lillicrap, and Gregory
  Wayne.
\newblock Experience replay for continual learning.
\newblock {\em Advances in Neural Information Processing Systems}, 32, 2019.

\bibitem{sariyildiz2023improving}
Mert~Bulent Sariyildiz, Yannis Kalantidis, Karteek Alahari, and Diane Larlus.
\newblock No reason for no supervision: Improved generalization in supervised
  models.
\newblock In {\em International Conference on Learning Representations}, 2023.

\bibitem{serra2018overcoming}
Joan Serra, Didac Suris, Marius Miron, and Alexandros Karatzoglou.
\newblock Overcoming catastrophic forgetting with hard attention to the task.
\newblock In {\em International Conference on Machine Learning}, 2018.

\bibitem{shi2021continual}
Yujun Shi, Li~Yuan, Yunpeng Chen, and Jiashi Feng.
\newblock Continual learning via bit-level information preserving.
\newblock In {\em Proceedings of the IEEE/CVF Conference on Computer Vision and
  Pattern Recognition}, pages 16674--16683, 2021.

\bibitem{tinyIM}
Stanford.
\newblock Tiny imagenet challenge, cs231n course., CS231N.

\bibitem{tang2021layerwise}
Shixiang Tang, Dapeng Chen, Jinguo Zhu, Shijie Yu, and Wanli Ouyang.
\newblock Layerwise optimization by gradient decomposition for continual
  learning.
\newblock In {\em Proceedings of the IEEE/CVF Conference on Computer Vision and
  Pattern Recognition}, pages 9634--9643, 2021.

\bibitem{wang2021ordisco}
Liyuan Wang, Kuo Yang, Chongxuan Li, Lanqing Hong, Zhenguo Li, and Jun Zhu.
\newblock Ordisco: Effective and efficient usage of incremental unlabeled data
  for semi-supervised continual learning.
\newblock In {\em Proceedings of the IEEE/CVF Conference on Computer Vision and
  Pattern Recognition}, pages 5383--5392, 2021.

\bibitem{wu2019large}
Yue Wu, Yinpeng Chen, Lijuan Wang, Yuancheng Ye, Zicheng Liu, Yandong Guo, and
  Yun Fu.
\newblock Large scale incremental learning.
\newblock In {\em Conference on Computer Vision and Pattern Recognition}, 2019.

\bibitem{xiao2021should}
Tete Xiao, Xiaolong Wang, Alexei~A Efros, and Trevor Darrell.
\newblock What should not be contrastive in contrastive learning.
\newblock {\em ICLR}, 2021.

\bibitem{yan2021dynamically}
Shipeng Yan, Jiangwei Xie, and Xuming He.
\newblock Der: Dynamically expandable representation for class incremental
  learning.
\newblock In {\em Proceedings of the IEEE/CVF Conference on Computer Vision and
  Pattern Recognition}, pages 3014--3023, 2021.

\bibitem{yu2020semantic}
Lu~Yu, { Twardowski, Bartlomiej}, { Liu, Xialei}, Luis Herranz, Kai Wang,
  Shangling Jui, and Joost van~de Weijer.
\newblock Semantic drift compensation for class-incremental learning.
\newblock In {\em Proceedings of the IEEE/CVF Conference on Computer Vision and
  Pattern Recognition}, pages 6982--6991, 2020.

\bibitem{zbontar2021barlow}
Jure Zbontar, Li~Jing, Ishan Misra, Yann LeCun, and St{\'e}phane Deny.
\newblock Barlow twins: Self-supervised learning via redundancy reduction.
\newblock {\em arXiv preprint arXiv:2103.03230}, 2021.

\bibitem{zenke2017continual}
Friedemann Zenke, Ben Poole, and Surya Ganguli.
\newblock Continual learning through synaptic intelligence.
\newblock In {\em International Conference on Machine Learning}, 2017.

\bibitem{10.1162/neco_a_01430}
Chen Zeno, Itay Golan, Elad Hoffer, and Daniel Soudry.
\newblock {Task-Agnostic Continual Learning Using Online Variational Bayes With
  Fixed-Point Updates}.
\newblock {\em Neural Computation}, 33(11):3139--3177, 10 2021.

\bibitem{zhai2021hyper}
Mengyao Zhai, Lei Chen, and Greg Mori.
\newblock Hyper-lifelonggan: Scalable lifelong learning for image conditioned
  generation.
\newblock In {\em Proceedings of the IEEE/CVF Conference on Computer Vision and
  Pattern Recognition}, pages 2246--2255, 2021.

\bibitem{zhang2020class}
Junting Zhang, Jie Zhang, Shalini Ghosh, Dawei Li, Serafettin Tasci, Larry
  Heck, Heming Zhang, and C-C~Jay Kuo.
\newblock Class-incremental learning via deep model consolidation.
\newblock In {\em Winter Conf. on App. of Computer Vision}, 2020.

\bibitem{zhu2021prototype}
Fei Zhu, Xu-Yao Zhang, Chuang Wang, Fei Yin, and Cheng-Lin Liu.
\newblock Prototype augmentation and self-supervision for incremental learning.
\newblock In {\em Proceedings of the IEEE/CVF Conference on Computer Vision and
  Pattern Recognition}, pages 5871--5880, 2021.

\bibitem{zini2023planckian}
Simone Zini, Alex Gomez-Villa, Marco Buzzelli, Bart{\l}omiej Twardowski,
  Andrew~D Bagdanov, and Joost van~de Weijer.
\newblock Planckian jitter: countering the color-crippling effects of color
  jitter on self-supervised training.
\newblock {\em ICLR}, 2023.

\end{thebibliography}

\clearpage

\appendix
\section{Appendix A}
\subsection{Other possible expert initializations}
\label{app:expert_init}

 We investigated two additional initialization regimes for the expert network:
 
\emph{ScratchOP}  - which is a trivial way to initialize $g_\phi^{t+1}$ with random weights. This initialization is fast and completely clean of any bias (either good or bad) of previous data. The main drawback is that previous knowledge could help $g_\phi^{t+1}$ to learn better representation. There is no knowledge transfer to the new expert.

\emph{FtOP} -  which begin the training of the next task using the expert from the current one. That is a reasonable assumption if the distributions of consecutive tasks are similar. This initialization copies the weights of $g_\phi^{t}$ into $g_\phi^{t+1}$, allowing a good initialization point for the training of the task $t+1$. This approach has two disadvantages: first, the knowledge of $g_\phi^{t}$ has a lot of recency bias ($g_\phi^{t}$ has already forgotten knowledge from tasks $t-1$); second, the architectures of $g_\phi^{t}$ and $g_\phi^{t+1}$ need to be homogenous to perform the copy operation.

Table \ref{tab:cifar100_appendix} show results of ScratchOP and FtOP expert initialization for POCON in CIFAR-100. The results for CaSSLe and PFR are recall once again here for an easy comparison.

\begin{table}[h!]
\centering
\caption{Accuracy of a linear evaluation on split CIFAR-100 for different number of tasks.}
\resizebox{\linewidth}{!}{
\begin{tabular}{l|lllll}
\midrule
\multicolumn{6}{c}{CIFAR-100 (32x32)}                                                                                                \\
\midrule
\textbf{Method}  & \textbf{4 tasks} & \textbf{10 tasks} & \textbf{20 tasks} & 5\textbf{0 tasks} & \textbf{100 tasks} \\
\midrule\midrule

FT  & 54.8  & 50.94  & 44.95  & 38.0   & 27.0         \\
CaSSLe& 59.80 &52.5 &49.6 & 45.3 & 42.10         \\

PFR& 59.70  & 54.33  & 44.80  & 46.5  & 43.30         \\

\midrule
\multicolumn{6}{c}{POCON}                                    \\
\midrule
ScratchOP& 57.64 & 46.87 & 39.69 &30.42 & 30.50         \\

FtOP&62.0 &58.13 &55.57 &48.64 &  47.06  \\
\midrule
Joint  & \multicolumn{5}{c}{65.4}        \\
\midrule
\end{tabular}
}
\label{tab:cifar100_appendix}
\end{table}

\subsection{Data partition for task-free setting experiment}

\begin{figure}[tb]
  \centering
  \includegraphics[width=0.4\textwidth]{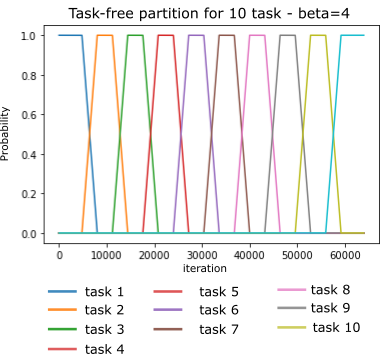}
  \caption{Sample data partition for 10 task and beta 4}
  \label{fig:partition}
\end{figure}

Fig.~\ref{fig:partition} depicts a data partition for $10$ tasks and beta 4 based on~\cite{10.1162/neco_a_01430}. The y-axis denotes the probability of selecting a sample in an iteration. Hence, the batches of data will contain a mixture of tasks data proportional to the probability of taking a sample for each class (beta$=4$ produces a mixture of two tasks). Fig.~\ref{fig:partition} shows where the mixed batches are taken, for instance at $8000$ iteration for this particular instance. In such created setting, there are no explicit tasks' boundaries that could be used in the continual training session to initiate additonal step for training a new task.

\end{document}